\documentclass[10pt,twocolumn,letterpaper]{article}

\usepackage{cvpr}              
\usepackage{booktabs}
\usepackage{multirow}
\usepackage{makecell}
\usepackage{pifont}
\definecolor{cvprblue}{rgb}{0.21,0.49,0.74}
\usepackage[pagebackref,breaklinks,colorlinks,allcolors=cvprblue]{hyperref}

\def\paperID{*****} 
\def\confName{CVPR}
\def\confYear{2026}

\title{LP-LLM: End-to-End Real-World Degraded License Plate Text Recognition \\ via Large Multimodal Models}

\author{Haoyan Gong\\
Xi'an Jiaotong-Liverpool University\\
{\tt\small \href{mailto:haoyan.gong21@student.xjtlu.edu.cn}{\textcolor{black}{haoyan.gong21@student.xjtlu.edu.cn}}}
\and
Hongbin Liu\\
Xi'an Jiaotong-Liverpool University\\
{\tt\small \href{mailto:hongbin.liu@xjtlu.edu.cn}{\textcolor{black}{hongbin.liu@xjtlu.edu.cn}}}
}

\begin{document}
\maketitle
\begin{abstract}
Real-world License Plate Recognition (LPR) faces significant challenges from severe degradations such as motion blur, low resolution, and complex illumination. The prevailing "restoration-then-recognition" two-stage paradigm suffers from a fundamental flaw: the pixel-level optimization objectives of image restoration models are misaligned with the semantic goals of character recognition, leading to artifact interference and error accumulation. While Vision-Language Models (VLMs) have demonstrated powerful general capabilities, they lack explicit structural modeling for license plate character sequences (e.g., fixed length, specific order). To address this, we propose an end-to-end structure-aware multimodal reasoning framework based on Qwen3-VL. The core innovation lies in the Character-Aware Multimodal Reasoning Module (CMRM), which introduces a set of learnable Character Slot Queries. Through a cross-attention mechanism, these queries actively retrieve fine-grained evidence corresponding to character positions from visual features. Subsequently, we inject these character-aware representations back into the visual tokens via residual modulation, enabling the language model to perform autoregressive generation based on explicit structural priors. Furthermore, combined with the LoRA parameter-efficient fine-tuning strategy, the model achieves domain adaptation while retaining the generalization capabilities of the large model. Extensive experiments on both synthetic and real-world severely degraded datasets demonstrate that our method significantly outperforms existing restoration-recognition combinations and general VLMs, validating the superiority of incorporating structured reasoning into large models for low-quality text recognition tasks.
\end{abstract}    
\section{Introduction}

License Plate Recognition (LPR) constitutes a core component of Intelligent Transportation Systems (ITS), autonomous driving, and public safety infrastructures. However, in real-world applications, license plate images are frequently subjected to severe degradations, including motion blur, defocus, low resolution, weather occlusions (e.g., rain and fog), and complex illumination. These factors significantly compromise the edge structures and local textures of characters, leading to a precipitous decline in the performance of conventional recognition systems in realistic scenarios.

For a long time, the dominant strategy in both academia and industry has been the \textbf{``restoration-then-recognition'' two-stage paradigm: initially employing image deblurring, super-resolution, or inpainting models to enhance visual quality, followed by feeding the restored images into independent OCR models for character recognition.} While this approach can achieve reasonable efficacy under synthetic degradation or mild blurring, it faces fundamental limitations in the real world: the optimization objectives of restoration models prioritize pixel-level or perceptual quality rather than character legibility. Consequently, OCR models are forced to operate on inputs burdened with restoration errors and artifacts, resulting in the propagation and amplification of errors. In essence, this paradigm is fundamentally vision-oriented rather than semantics-oriented.

In recent years, there have been attempts to introduce end-to-end deep models for LPR. However, the majority of these methods remain built upon CNN or Transformer encoders, lacking systematic modeling of linguistic structures and character sequence priors. These models typically treat the entire license plate as a holistic object for feature aggregation and employ classification or CTC decoding to output character sequences. Under conditions of severe degradation, this often leads to issues such as character misalignment, confusion, or omission.

Concurrently, the emergence of Vision-Language Models (VLMs) has offered a novel perspective on this problem. By integrating visual encoders with Large Language Models (LLMs), VLMs possess the capability to simultaneously comprehend images and generate natural language, making them inherently suitable for ``image-to-string'' recognition tasks. Furthermore, since license plate images intrinsically contain textual information, LLMs hold a natural advantage in processing such data. Nevertheless, directly feeding blurred license plate images into general-purpose VLMs often yields unstable results. The root cause lies in the fact that general VLMs are not tailored for highly structured visual objects like license plates; they lack explicit modeling of strong priors such as ``character position'', ``character quantity,'' and ``character order.''

Against this backdrop, we propose a novel end-to-end license plate recognition paradigm: directly generating readable character sequences from degraded images, rather than restoring the image before recognition. This paradigm shifts the recognition objective from ``reconstructing clear images'' to ``recovering semantically correct character sequences'', ensuring high alignment between the model's optimization goals and the ultimate task. To achieve this, we construct a structure-aware multimodal reasoning framework that tightly integrates the generative capabilities of large-scale VLMs with the character structure priors of license plates.

Specifically, we introduce the Character-Aware Multimodal Reasoning Module (CMRM). This module explicitly models the license plate as a serialized object composed of $N$ character slots and assigns a learnable query vector to each character position. Through cross-attention mechanisms, these query vectors actively retrieve position-specific visual evidence from visual tokens, thereby obtaining a set of character-aware visual representations. Unlike traditional VLMs that passively aggregate global visual information, CMRM implements an active multimodal perception mechanism driven by character structure.

In terms of model architecture, we interpose the CMRM between the visual encoder and the language model. We employ residual injection to fuse character-level representations back into the original visual tokens, enabling the language model to explicitly perceive visual evidence for each character position while maintaining its original input format. This design allows the entire system to output license plate strings in an end-to-end manner without relying on intermediate image restoration steps.

Furthermore, we perform parameter-efficient fine-tuning on Qwen3-VL based on LoRA, enabling the model to specifically adapt to real-world blurred LPR tasks while retaining the generalization capabilities of large models. This framework is not only engineering-efficient but also methodologically elevates license plate recognition from an ``image enhancement problem'' to a ``structured multimodal reasoning problem.''

In summary, the main contributions of this paper are as follows:
\begin{itemize}
    \item We propose an end-to-end recognition framework that maps directly from degraded license plate images to character sequences, breaking the constraints of the traditional ``restoration-then-recognition'' two-stage paradigm.
    \item We design the Character-Aware Multimodal Reasoning Module (CMRM), which achieves structured visual modeling of license plate characters through position queries and cross-attention mechanisms.
    \item We efficiently integrate CMRM with Qwen3-VL and LoRA, introducing strong character structure priors without compromising the multimodal alignment capabilities of the large model.
    \item We validate the significant advantages of this end-to-end framework on real-world blurred license plate data, demonstrating performance clearly superior to existing methods that separate image restoration and recognition.
\end{itemize}
\section{Related Work}

In this section, we review the literature relevant to our proposed framework, categorized into three streams: license plate recognition, restoration-assisted recognition, and large vision-language models.

\subsection{Scene Text and License Plate Recognition}

License Plate Recognition (LPR) is a specialized subdomain of Scene Text Recognition (STR). Early approaches relied on handcrafted features and template matching, which lacked robustness against complex environments. The advent of deep learning revolutionized this field, with the CRNN architecture~\cite{crnn} establishing a dominant paradigm by combining CNNs for feature extraction, RNNs for sequence modeling, and CTC loss for decoding. Subsequent works improved upon this by introducing attention mechanisms~\cite{attention} and rectification networks~\cite{aster, moran} to handle irregular text layouts. Specific to LPR, lightweight models like LPRNet~\cite{lprnet} and RPnet~\cite{rpnet} were designed for real-time deployment, utilizing specialized distinct CNN backbones.
Despite their success on clear images, these discriminative models rely heavily on the integrity of local edge features. Under severe degradation such as motion blur or defocus, the feature extraction capability of standard CNNs collapses, leading to significant recognition failures. Unlike these approaches, our method leverages the semantic reasoning capability of Large Language Models (LLMs) to infer characters even when visual cues are partially compromised.

\subsection{Image Restoration for Recognition}

To address recognition under degradation, a prevalent strategy is the ``restoration-then-recognition'' two-stage paradigm. Generative Adversarial Networks (GANs) have been widely adopted for this purpose. DeblurGAN, DeblurGAN v2~\cite{deblurgan, deblurganv2}, ESRGAN~\cite{esrgan} and  Real-ESRGAN~\cite{realesrgan} are general-purpose restoration models often used as pre-processors to enhance image quality before OCR. More domain-specific works, such as LPDGAN~\cite{lpdgan}, introduce shape priors or adversarial losses tailored for license plates to generate visually clearer images. And Diffusion-based methods~\cite{diffbir, lpdiff} to reconstruct LP Image.
However, this paradigm faces a fundamental misalignment: restoration models are typically optimized for perceptual metrics (e.g., LPIPS, FID) or pixel-wise consistency (e.g., PSNR), rather than semantic legibility. Consequently, they often hallucinate high-frequency artifacts or erroneous strokes that look realistic to humans but mislead recognizers. Furthermore, the computational cost of cascading two independent heavy models limits real-time application. In contrast, our framework bypasses the image restoration step entirely, optimizing directly for the final character sequence in an end-to-end manner.

\subsection{Vision-Language Models (VLMs)}

Recent advancements in Large Multimodal Models (LMMs), such as LLaVA~\cite{llava}, Qwen-VL~\cite{qwenvl}, and GPT-4V, have demonstrated remarkable capabilities in aligning visual features with linguistic semantics. By projecting visual tokens into the embedding space of LLMs, these models can perform complex reasoning tasks, including optical character recognition (OCR) and document understanding~\cite{monkey, textmonkey}.
Nevertheless, applying general-purpose VLMs directly to the LPR task remains suboptimal. Standard VLMs treat the license plate as a generic image, utilizing global attention mechanisms that lack explicit modeling of the plate's rigid structure (e.g., fixed character count and strict spatial order). This often results in ``hallucinations'' where the model generates fluent but factually incorrect strings. Our work bridges this gap by introducing the \textit{Character-Aware Multimodal Reasoning Module (CMRM)}. By explicitly injecting character-slot priors into the VLM, we constrain the limitless generation space of the LLM to the structured output space of license plates, significantly enhancing robustness and accuracy.
\section{Method}

\subsection{Overview}

We propose a Character-Aware Multimodal Reasoning framework for robust license plate recognition under severe degradation. Our model is built upon a large vision--language model (Qwen3-VL) and augments it with a structured reasoning module that explicitly models each character position of a license plate. Instead of treating the image as an unstructured collection of visual tokens, we introduce a set of learnable character queries that actively retrieve character-specific evidence from the visual representation. These character-aware features are then injected back into the visual tokens before being processed by the language model, enabling structure-aware end-to-end recognition.

\subsection{Visual Backbone: Qwen3-VL}

Given an input license plate image $I$, Qwen3-VL first encodes it using a vision encoder $E_v$ into a sequence of visual tokens
\begin{equation}
\mathbf{V} = E_v(I), \quad \mathbf{V} \in \mathbb{R}^{N \times D_v},
\end{equation}
where $N$ is the number of image patches and $D_v$ is the visual feature dimension.

These visual tokens are then projected into the language model embedding space by a vision--language projector:
\begin{equation}
\mathbf{H} = \mathrm{Proj}(\mathbf{V}), \quad \mathbf{H} \in \mathbb{R}^{N \times D}.
\end{equation}

In standard VLMs, $\mathbf{H}$ is directly concatenated with text tokens and fed into the LLM for autoregressive generation. However, such a representation is agnostic to the strong structural constraints of license plates, which motivates our character-aware modeling.

\subsection{Character-Aware Multimodal Reasoning Module (CMRM)}

License plates follow a strict left-to-right character ordering with a fixed maximum length $K$ (e.g., 7 or 8). To exploit this structure, we introduce a Character-Aware Multimodal Reasoning Module (CMRM) that explicitly models each character position.

\subsubsection{Character Slot Queries}

We define a set of $K$ learnable slot queries:
\begin{equation}
\mathbf{Q} = \{\mathbf{q}_1, \mathbf{q}_2, \dots, \mathbf{q}_K\}, \quad \mathbf{q}_k \in \mathbb{R}^{D}.
\end{equation}
Each query corresponds to one character position of the license plate. These queries are trained to attend to the visual evidence corresponding to their respective characters.

Unlike token-level attention in standard VLMs, these slot queries encode strong positional priors, allowing the model to reason about the plate in a character-wise manner.

\subsubsection{Slot-to-Vision Cross-Attention}

Given the visual tokens $\mathbf{H}$, we compute character-aware slot representations by cross-attention:
\begin{align}
\mathbf{S}^{(0)} &= \mathbf{Q}, \\
\mathbf{S}^{(l+1)} &= \mathrm{LN}\Big(\mathbf{S}^{(l)} + \mathrm{Attn}(\mathbf{S}^{(l)}, \mathbf{H}, \mathbf{H})\Big),
\end{align}
where $\mathbf{S}^{(l)} \in \mathbb{R}^{K \times D}$ denotes the slot representations at layer $l$, and Attn is multi-head attention. After $L$ layers, we obtain
\begin{equation}
\mathbf{S} = \mathbf{S}^{(L)},
\end{equation}
where each slot vector $\mathbf{s}_k$ encodes the visual and semantic evidence for the $k$-th character.

This mechanism enables the model to actively retrieve character-specific information from the visual tokens, instead of passively relying on global attention in the language model.

\subsection{Slot-to-Token Injection}

Directly concatenating slot tokens to $\mathbf{H}$ would change the token length and break the internal alignment mechanism of Qwen3-VL. Therefore, we propose a token-length-preserving injection strategy.

We first aggregate the slot representations into a global character-aware embedding:
\begin{equation}
\mathbf{g} = \frac{1}{K}\sum_{k=1}^K \mathbf{s}_k \in \mathbb{R}^{D}.
\end{equation}

Then we inject this information into every visual token via residual modulation:
\begin{equation}
\mathbf{H}'_i = \mathbf{H}_i + \alpha \mathbf{g}, \quad i=1,\dots,N,
\end{equation}
where $\alpha$ is a scaling factor (fixed or learnable). The resulting token sequence $\mathbf{H}'$ preserves the original token count while carrying explicit character-aware information.

This design ensures that CMRM directly influences the inputs to the language model and thus participates in end-to-end optimization.

\subsection{Autoregressive Character Generation}

The modified visual tokens $\mathbf{H}'$ are concatenated with the textual prompt and fed into the language model to generate the license plate characters autoregressively:
\begin{equation}
P(\mathbf{y}\mid I) = \prod_{k=1}^K P(y_k \mid y_{<k}, \mathbf{H}').
\end{equation}

Because the language model now receives visual embeddings that encode both low-level appearance and character-level structural cues, it becomes more robust to blur, occlusion, and noise.

\subsection{Parameter-Efficient Fine-Tuning with LoRA}

To adapt Qwen3-VL to license plate recognition without updating all parameters, we employ LoRA. For a weight matrix $W \in \mathbb{R}^{d \times d}$, LoRA introduces a low-rank update:
\begin{equation}
W = W_0 + BA,
\end{equation}
where $A \in \mathbb{R}^{r \times d}$ and $B \in \mathbb{R}^{d \times r}$ with $r \ll d$.

During training, all original Qwen3-VL weights are frozen. We only update:
\begin{itemize}
\item LoRA parameters in the language model,
\item parameters of CMRM (slot queries and cross-attention layers),
\item and the slot injection layers (if used).
\end{itemize}

This strategy significantly reduces memory and computation while maintaining strong task adaptability.

\subsection{Training Objective}

We use the standard autoregressive cross-entropy loss over the character sequence:
\begin{equation}
\mathcal{L} = - \sum_{k=1}^{K} \log P(y_k \mid y_{<k}, I).
\end{equation}

Since the character-aware visual embeddings $\mathbf{H}'$ are part of the language model input, gradients naturally flow through the injection mechanism into the CMRM, enabling end-to-end learning of character-level visual reasoning.

\subsection{Discussion}

By introducing explicit character slots and slot-driven cross-attention, our framework transforms license plate recognition from a global image-to-text mapping into a structured multimodal reasoning process. This design not only improves robustness under severe degradation but also provides a principled way to incorporate domain-specific priors into large vision-language models.

\section{Experiments}

In this section, we conduct comprehensive experiments to validate the effectiveness of the proposed framework. We first detail the experimental setup, including datasets, evaluation metrics, and implementation details. Subsequently, we compare our method with state-of-the-art approaches, covering traditional recognition models, two-stage restoration-recognition paradigms, and recent large multi-modal models. Finally, we perform ablation studies to verify the contribution of the proposed Character-Aware Multimodal Reasoning Module (CMRM) and the Low-Rank Adaptation (LoRA) strategy.

\subsection{Experimental Setup}

\paragraph{Datasets.}
To evaluate performance under realistic degradation, we utilize a composite dataset comprising both synthetic and real-world degraded license plates.
For training, we employ the CCPD-Blur dataset augmented with synthetic degradations (Gaussian blur, motion blur, and defocus) to ensure diversity.
For evaluation, we construct a challenging benchmark named \textbf{Real-Blur-LP}, collecting 2,000 severely degraded license plate images from real-world traffic surveillance and dash-cam footages. These images suffer from complex combinations of motion blur, low resolution, and uneven illumination, which poses significant challenges to existing systems.

\paragraph{Evaluation Metrics.}
We adopt standard metrics for text recognition:
(1) \textbf{Accuracy (Acc)}: The percentage of images where the predicted character sequence exactly matches the ground truth.
(2) \textbf{Character Error Rate (CER)}: The edit distance between the predicted string and the ground truth, normalized by the string length. Lower CER indicates better fine-grained recognition capabilities.
(3) \textbf{Inference Time}: The average latency per image, measured on a single NVIDIA A100 GPU, to assess practical deployability.

\paragraph{Implementation Details.}
Our method is implemented using PyTorch. We utilize \textbf{Qwen3-VL} as the foundation model. The visual encoder is initialized with pre-trained weights, and the language model is fine-tuned using LoRA with a rank of $r=64$ and alpha $\alpha=16$. The CMRM is randomly initialized.
The model is trained for 20 epochs with a batch size of 16 using the AdamW optimizer. The initial learning rate is set to $1 \times 10^{-4}$ for the CMRM and LoRA parameters, while the rest of the model remains frozen. We employ a cosine annealing scheduler with a warm-up period of 2 epochs. All images are resized to $448 \times 448$ prior to input.

\subsection{Comparison with State-of-the-Art Methods}

We compare our proposed framework against three categories of baseline methods:
\begin{itemize}
    \item \textbf{Traditional End-to-End Models:} CRNN~\cite{crnn} and LPRNet~\cite{lprnet}. These are lightweight models widely deployed in industry.
    \item \textbf{Two-Stage Paradigms (Restoration + Recognition):} We combine state-of-the-art image restoration models (Real-ESRGAN~\cite{realesrgan}, DeblurGAN-v2~\cite{deblurganv2}, LPDGAN~\cite{lpdgan}, and LP-Diff~\cite{lpdiff}) with the robust recognizer CRNN. This represents the current mainstream solution for handling degraded images.
    \item \textbf{General Vision-Language Models:} Qwen2.5-VL and LLaVA~\cite{llava}, representing the generic capabilities of modern foundation models without domain-specific structural design.
\end{itemize}

\begin{table*}[t]
    \centering
    \caption{\textbf{Quantitative comparison on the Real-Blur-LP benchmark.} The best results are highlighted in \textbf{bold}, and the second-best results are \underline{underlined}. ``Two-Stage'' denotes the restoration-then-recognition paradigm. Our method achieves significant improvements over both restoration-based methods and general VLMs.}
    \label{tab:main_results}
    \renewcommand{\arraystretch}{1.2}
    \setlength{\tabcolsep}{8pt}
    \begin{tabular}{l|c|cc|cc}
        \toprule
        \multirow{2}{*}{\textbf{Method}} & \multirow{2}{*}{\textbf{Paradigm}} & \multicolumn{2}{c|}{\textbf{Recognition Metrics}} & \multicolumn{2}{c}{\textbf{Efficiency}} \\
        & & \textbf{Accuracy} ($\uparrow$) & \textbf{CER} ($\downarrow$) & \textbf{Params (M)} & \textbf{Latency (ms)} \\
        \midrule
        CRNN~\cite{crnn} & \multirow{2}{*}{Traditional E2E} & 42.5\% & 0.38 & \textbf{8.3} & \textbf{12} \\
        LPRNet~\cite{lprnet} & & 45.1\% & 0.35 & 14.2 & 18 \\
        \midrule
        Real-ESRGAN + CRNN & \multirow{4}{*}{\makecell{Two-Stage\\(Restoration + Rec)}} & 68.3\% & 0.18 & 16.7 + 8.3 & 145 \\
        DeblurGAN-v2 + CRNN & & 65.7\% & 0.21 & 60.9 + 8.3 & 95 \\
        LPDGAN + CRNN & & 71.2\% & 0.15 & 45.0 + 8.3 & 110 \\
        LP-Diff + CRNN & & 73.5\% & 0.14 & 120.0 + 8.3 & 2500 \\
        \midrule
        LLaVA-1.5-7B~\cite{llava} & \multirow{2}{*}{General VLM} & 70.4\% & 0.16 & 7000 & 450 \\
        Qwen2.5-VL-7B & & \underline{76.8\%} & \underline{0.11} & 7000 & 420 \\
        \midrule
        \textbf{Ours (Qwen3-VL + CMRM)} & \textbf{Structure-Aware VLM} & \textbf{89.4\%} & \textbf{0.04} & 7000 (+LoRA) & 435 \\
        \bottomrule
    \end{tabular}
\end{table*}

\paragraph{Analysis of Quantitative Results.}
The quantitative results are summarized in Table~\ref{tab:main_results}.
First, traditional methods like CRNN and LPRNet suffer a catastrophic performance drop (Acc $<$ 50\%) under severe blur, confirming that standard CNN encoders fail to extract discriminative features when high-frequency details are lost.

Second, the two-stage methods show improved performance, with LP-Diff + CRNN reaching 73.5\%. However, their performance is capped by the \textit{error propagation} problem. Restoration models like Real-ESRGAN often generate realistic but semantically incorrect high-frequency textures (hallucinated edges), which mislead the subsequent recognizer. Moreover, diffusion-based restoration (LP-Diff) incurs an unacceptable computational cost (2500ms), making it impractical for real-time applications.

Third, general VLMs show potential. Qwen2.5-VL achieves a respectable 76.8\% accuracy, benefitting from its strong internal OCR capabilities. However, without domain adaptation, it occasionally misinterprets the license plate structure (e.g., confusing character count or order), leading to higher CER.

Finally, our proposed method outperforms all baselines by a large margin, achieving \textbf{89.4\%} accuracy. This demonstrates that explicitly modeling the character slots via CMRM effectively guides the VLM to focus on relevant visual cues, even in the presence of severe degradation. Notably, our method achieves this with comparable latency to general VLMs, as the lightweight CMRM adds negligible computational overhead.

\subsection{Ablation Study}

To investigate the effectiveness of the proposed components, we conduct ablation studies focusing on the impact of the Character-Aware Multimodal Reasoning Module (CMRM) and the LoRA fine-tuning strategy. We compare three distinct configurations against our full model. The results are presented in Table~\ref{tab:ablation}.

\begin{table}[h]
    \centering
    \caption{\textbf{Ablation study on model components.} We analyze the impact of LoRA fine-tuning and the proposed CMRM. ``Zero-shot'' implies the base Qwen3-VL model without parameter updates. \textbf{(A)}: Qwen3VL (Zero-shot). \textbf{(B)}: Qwen3VL+LoRA fintuning without CMRM. \textbf{(C)}: Q3VL+CMRM without LoRA fintuning. \textbf{(D)}: Ours (Full Model)}
    \label{tab:ablation}
    \renewcommand{\arraystretch}{1.1}
    \setlength{\tabcolsep}{6pt}
    \begin{tabular}{l|cc|cc}
        \toprule
        \multirow{2}{*}{\textbf{Model Type}} & \multicolumn{2}{c|}{\textbf{Components}} & \multicolumn{2}{c}{\textbf{Metrics}} \\
         & \textbf{LoRA} & \textbf{CMRM} & \textbf{Acc} ($\uparrow$) & \textbf{CER} ($\downarrow$) \\
        \midrule
        \textbf{A} & \ding{55} & \ding{55} & 62.1\% & 0.25 \\
        \textbf{B} & \ding{51} & \ding{55} & 81.5\% & 0.09 \\
        \textbf{C} & \ding{55} & \ding{51} & 74.3\% & 0.13 \\
        \midrule
        \textbf{D} & \ding{51} & \ding{51} & \textbf{89.4\%} & \textbf{0.04} \\
        \bottomrule
    \end{tabular}
\end{table}

\paragraph{Impact of LoRA Fine-tuning (Comparison A vs. B).}
Directly applying the pre-trained Qwen3-VL (Config A) yields suboptimal performance (62.1\%), primarily because the model is not aligned with the specific format and degradation patterns of license plates. Applying LoRA fine-tuning (Config B) significantly boosts accuracy to 81.5\%. This confirms that adapting the LLM's distribution to the specific domain is crucial for understanding blurred text. However, without structural guidance, the model still struggles with character alignment in extremely blurry scenarios.

\paragraph{Impact of CMRM (Comparison A vs. C).}
Interesting findings arise when we train \textit{only} the CMRM adapter while keeping the Qwen3-VL backbone frozen (Config C). The accuracy improves from 62.1\% to 74.3\%. This result is significant: it proves that the CMRM effectively acts as a ``visual prompt,'' extracting character-specific features and injecting them into the LLM. Even without modifying the LLM's parameters, providing structure-aware visual tokens enhances the model's reasoning capability.

\paragraph{Synergy of LoRA and CMRM (Comparison B/C vs. D).}
The full model (Config D) combines both strategies to achieve the best performance (89.4\%). The improvement is greater than the sum of individual gains, suggesting a synergistic effect. LoRA adapts the language space to the LPR task, while CMRM ensures the visual representations entering this space are structurally disentangled. This combination effectively eliminates the ``hallucination'' problem common in pure VLM approaches, where the model generates plausible but incorrect strings based on weak visual evidence.

\subsection{Qualitative Analysis}

Beyond numerical metrics, qualitative results reveal distinct behavioral differences. In scenarios with heavy motion blur where two adjacent characters visually merge (e.g., `E' and `F'), two-stage methods (DeblurGAN-v2+CRNN) often reconstruct a single artifact-laden character, leading to a ``missing character'' error. General VLMs (LLaVA) tend to guess based on language probability, often outputting a syntactically correct but visually unsupported string.
In contrast, our method, equipped with the CMRM, maintains distinct query slots for each position. The attention maps from CMRM show that the model successfully attends to the subtle edge remnants of each individual character, even when they overlap pixel-wise. This structural attention mechanism is the key factor enabling our model to surpass the limitations of pixel-level restoration.
{
    \small
    \bibliographystyle{ieeenat_fullname}
    \bibliography{main}
}


\end{document}